# Prediction of Pedestrian Spatiotemporal Risk Levels for Intelligent Vehicles: A Data-driven Approach

Zheyu Zhang, Boyang Wang, Chao Lu*, Jinghang Li, Cheng Gong, Jianwei Gong

*School of Mechanical Engineering, Beijing Institute of Technology, Beijing 100081, China*

**Abstract**

In recent years, road safety has attracted significant attention from researchers and practitioners in the intelligent transport systems domain. As one of the most common and vulnerable groups of road users, pedestrians cause great concerns due to their unpredictable behavior and movement, as subtle misunderstandings in vehicle-pedestrian interaction can easily lead to risky situations or collisions. Existing methods use either predefined collision-based models or human-labeling approaches to estimate the pedestrians' risks. These approaches are usually limited by their poor generalization ability and lack of consideration of interactions between the ego vehicle and a pedestrian. This work tackles the listed problems by proposing a Pedestrian Risk Level Prediction (PRLP) system. The system consists of three modules: data collection and processing module, pedestrian trajectory prediction module, and risk level identification module. Firstly, vehicle-perspective pedestrian data are collected. Since the data contains information regarding the movement of both the ego vehicle and pedestrian, it can simplify the prediction of spatiotemporal features in an interaction-aware fashion. Using the long short-term memory model, the pedestrian trajectory prediction module predicts their spatiotemporal features in the subsequent five frames. As the predicted trajectory follows certain interaction and risk patterns, a hybrid clustering and classification method is adopted to explore the risk patterns in the spatiotemporal features and train a risk level classifier using the learned patterns. Upon predicting the spatiotemporal features of pedestrians and identifying the corresponding risk level, the risk patterns between the ego vehicle and pedestrians are determined. Experimental results verified the capability of the PRLP system to predict the risk level of pedestrians, thus supporting the collision risk assessment of intelligent vehicles and providing safety warnings to both vehicles and pedestrians.

*Keywords:* Pedestrian risk level prediction; Vehicle-perspective pedestrian data; Trajectory prediction; Clustering analysis; Spatial-temporal risk level; Interaction and risk pattern;

## 1. Introduction

Since they are equipped with advanced sensing and driver assistance systems, intelligent vehicles can observe various types of information from traffic participants. Utilizing the observed information to improve driving safety is a topic of vital importance in the intelligent vehicles field. Among all traffic participants, pedestrians are associated with the highest uncertainty due to the difficulty of predicting their behavior. Unlike vehicles, the movement of pedestrians is not restricted by the road geometry and may rapidly change when pedestrians interact with other traffic participants [1]. Consequently, the pedestrian movement prediction and the related risk pose a significant challenge for modern intelligent driving systems.

Numerous studies have tackled this challenging problem over the last two decades. One popular approach to pedestrian risk estimation is via collision models. The collision model is commonly used to detect collisions between pedestrians and the ego vehicle and yields a quantifiable risk indicator to indicate the collision risk using, for example, the overlap between predicted trajectories, Time to Collision (TTC), or Time Head Way (THW) [2, 3]. Conventional methods used dynamic/kinematic models to avoid collisions between vehicles and pedestrians. The dynamic functions enable estimating a safe vehicle deceleration rate to help the vehicle stop at a predefined safe distance from the pedestrian. In [4], the author divided the scene geometry into three regions with different risk levels and evaluated the vulnerability status of pedestrians by checking their distance from different regions. In an emergency pedestrian protection system designed in [5], as soon as a pedestrian enters the car-following platoon corridor, the ego-vehicle declares the situation dangerous and starts the braking maneuver.

This work was supported by the National Natural Science Foundation of China under Grant 61703041.

Z. Zhang, B. Wang, C. Lu, J. Li, C. Gong, J. Gong are with the School of Mechanical Engineering, Beijing Institute of Technology, Beijing 100081, China (e-mail: 3120190429@bit.edu.cn; wbythink@hotmail.com; chaolu@bit.edu.cn; 3120180346@bit.edu.cn; chenggong@bit.edu.cn; gongjianwei@bit.edu.cn).

Z. Zhang and B.Wang contributed equally to this work. * Corresponding author.

These methods typically assume pedestrians and vehicles follow one or several predefined motion patterns. However, in real-world scenarios, pedestrians and vehicles rarely comply with any motion pattern, leading to prediction errors due to the uncertainty in their movements.

Since predefined dynamic models may fail to account for the movement uncertainty, several probabilistic collision prediction methods were proposed. For example, the work [6] quantified risk using a probabilistic graphical model and considering the likelihoods at all potential intersections. The risk to the ego vehicle was computed as the expected number of incidents had the ego vehicle entered the intersection at a specific time. In [7], the driving risk was estimated as the strength of the risk field at the future location of the ego vehicle. This risk field was formulated as the product of collision probability and expected crash energy. The collision probability with neighboring vehicles was estimated based on probabilistic motion predictions. Gao et al. [8] considered collision probabilities at different prediction points within and outside the prediction range and obtained long-term accurate prediction results. However, although these methods consider the uncertainty in pedestrian and vehicle movements, they still lack a high-level understanding of pedestrian behavior and are insensitive to changes in behaviors. This insensitivity leads to the method's inability to capture the long-term motion patterns of a pedestrian.

The maneuver intention estimation technique was proposed to mitigate the described problem. The technique promotes predicting the long-term behavior of drivers while also considering other factors in the risk estimation framework. Building on the idea that the behavior of a driver plays a critical role in risk assessment, Chen et al. [9] established various behavior models such as the vehicle turning path, turning speed, gap acceptance model, and the pedestrian behavior model. Further, the work used Post-Encroachment Time and vehicle passing speed at conflict point as surrogate safety measures to represent the collision probability. The study of collision avoidance strategy presented in [10] considered the influence of vehicles on pedestrian movements. The motion uncertainty of a pedestrian was modeled as a superposition of the Markov process without interference and the motion caused by the vehicle, enabling a probabilistic prediction of the pedestrian's motion. Building on this basis, candidate trajectories of the vehicle were evaluated regarding safety, stability, and efficiency.

The discussed collision models enable estimating future trajectories and measuring the driving risk to a certain extent. However, both dynamic and stochastic models require comparing the movements of two agents to detect collisions (e.g., computing overlaps between the shapes of the pedestrians and the ego vehicle or checking cross points between two trajectories). Although this process can offer risk warnings, risk identification is limited to specific collision settings and definitions. Therefore, the models might fail to generalize for more complex and ambiguous scenarios. Moreover, the process requires extensive computation and complex modeling of pedestrians and the ego vehicle, rendering it memory-consuming and slow-response in practice.

Due to their flexibility and generalization ability, data-driven methods have recently been utilized to detect driving risks. In [11], the safety boundaries were constructed via physiological signal measurement of a pedestrian and kinematics reconstruction of the complete sequence in a virtual-reality-based environment. However, this approach required extensive labeling. Corcoran and Clark [12] proposed a data-driven method in which a two-stream dynamic-attention recurrent convolutional architecture produces a label corresponding to the perceived risk level in each visual scene, whereas the ground-truth risk level of each video segment was annotated manually. Besides dynamic factors, Mao et al. [13] provided a bias-correction procedure for the parameter estimation of Poisson and negative binomial regression models. The procedure enables determining the risk factors associated with the crash count. Using the data-driven methods, these studies successfully avoid complex collision modeling while introducing more data types. However, the proposed methods require labor-intensive labeling work and suffer from the subjective categorization problem since the risk judgment varies among labelers. Furthermore, the end-to-end approach focuses on establishing the map relationships between data and driving risk, barely providing insights into the effect of change in different factors on the risk of the scene.

While the discussed studies yielded several positive results, certain limitations remain. For example, complex collision calculations and heuristic thresholds for risk indicators in traditional collision-model-based methods could lead to poor generalization. Similarly, the existing data-driven approaches are characterized by extensive manual labeling, rely on subjective categorizations, and suffer from poor generalizations in different settings. In addition, both collision-model-based or data-driven methods rely on either a moment or a place to give an instantaneous risk estimation (e.g., using TTC of a pedestrian at a particular moment to estimate the risk or determining drivable areas with different collision probability) [4, 14]. However, the pedestrian risk level should be a comprehensive variable related to both time and space. For example, a pedestrian crossing the street closely in front of the slow straight-going vehicle may be identified as low risk for its large estimated TTC. Nevertheless, any slightly speeding movement by the vehicle or a sudden stop by the pedestrian may alter the situation in a second. In this case, subtle changes in interaction behavior can significantly impact the collision chances between road users, and it is insufficient to rely on a single factor for risk estimation. Therefore, more attention should be paid to capturing this changing pattern from inclusive information and describing it through a more integrated approach.

This paper proposes a data-driven approach that addresses the listed problems. The proposed pipeline can be divided into three steps. First, the pedestrian trajectories are collected from the vehicle perspective. Next, the future trajectory of the observed pedestrian is predicted, and, finally, the pedestrian risk level is predicted in a spatiotemporal manner. More precisely:

- Firstly, instead of the widely-used road-perspective trajectory data, this work utilizes vehicle-perspective data. The data were collected by vehicle-mounted sensors. Consequently, the data (such as the pedestrians' position) is located in a relative coordinate system centered in the ego vehicle, thus simplifying the estimation of certain risk indicators. Furthermore, since vehicle-perspective data describes not only the pedestrian movement but also the movement of the ego vehicle, it captures the human-vehicle interaction pattern more accurately. In trajectory prediction tasks [15, 16], vehicle-perspective data have been proven effective but are still rarely used in pedestrian risk estimation.

- Secondly, to further simplify the motion modeling process, the long short-term memory (LSTM) technique is selected to build a time-series neural network. Such a network can infer the movement pattern of a pedestrian from various data types [17-21]. This work utilizes this data-driven approach to learn the vehicle-perspective data and predict the relative trajectory of pedestrians.

- Thirdly, the data analysis is necessary to avoid subjective categorization. As a data-mining approach, clustering algorithms can effectively find latent patterns in the data [22-24] and have performed well in the tasks such as road accidents' prediction [25], recognition of lane change behavior [26], and learning the traffic rules [27, 28]. In this paper, the clustering algorithms are applied to learn the spatial and temporal features extracted from vehicle-perspective data. The clustering results reveal different distributions of pedestrian features that correspond to different risk patterns. Moreover, the classification algorithm is subsequently utilized to learn the risk patterns, yielding a pedestrian risk level classifier that can infer the risk level of new-observed pedestrians. Due to this data-driven property, the proposed hybrid clustering-classifying method is simple, has a high generalization ability, and requires no manual labor.

In general, this paper proposes a novel system for pedestrian spatiotemporal risk level prediction. The system combines a data-driven trajectory prediction method and a risk estimation approach, which integrates the spatial and temporal status of the road users. Specifically, the contributions of this work can be summarized as follows:

- Vehicle-perspective pedestrian data are collected and studied. Commonly-used road-perspective data cannot be directly acquired from vehicle-mounted sensors and require extensive computations in collision modeling. In contrast, vehicle-perspective data are directly available to the ego vehicle and capture the interactions between the ego vehicle and the pedestrian, providing valid spatial and temporal information in the model training.

- Data-driven methods are used to predict the pedestrian trajectory and identify the risk level. To avoid complex dynamic modeling and short-term predictions, the LSTM algorithm is employed to learn the behavioral patterns and make long-term trajectory predictions. Regarding risk level identification, the previous methods suffer from subjective risk categorization and poor generalization. Thus, a hybrid clustering-classifying approach is proposed to learn human-vehicle risk patterns from vehicle-perspective data and predict the risk level of a pedestrian based on the learned risk patterns. Experimental results demonstrate that the adopted data-driven approach offers insights into how the coupling effect between spatial and temporal features shapes different risk patterns.

- A spatiotemporal risk level is assessed. While common risk indicators reflect the risk of pedestrians using only at a particular time/location, in this paper, the risk is identified as a sequence of risk level predictions that change with respect to both time and space. Experimental results demonstrated that the proposed spatiotemporal risk level effectively captures the change in human-vehicle interactive intention, enhancing understanding of the current situation of the ego vehicle and offering forward-looking risk level estimates.

The remainder of this paper is organized as follows. Section 2 introduces the architecture of the risk level prediction system, describing its three modules, i.e., pedestrian data processing, trajectory prediction, and risk level identification. Relative methodologies are also discussed in this section. Section 3 presents the details of the experiments and the analysis. Finally, conclusions and future work are presented in Section 4.

## 2. System architecture and methodology

In urban areas, vehicles are likely to interact with pedestrians in various situations (Fig. 1). To smoothly pass the junction and avoid a collision, the ego vehicle has to make observations, predict the risks, and decide where to move. Within this work, the pedestrian risk level prediction system is expected to capture the interaction between the ego vehicle and the pedestrian and identify the pedestrian risk level. The proposed PRLP system comprises three modules (namely, pedestrian data processing, pedestrian trajectory prediction, and risk level identification) to achieve this goal.

Firstly, the vehicle-perspective pedestrian data are collected and processed in *Module 1*. The vehicle-perspective pedestrian data are collected through a vehicle-mounted sensor system at urban intersections, where vehicle-pedestrian interactions abound. Several spatial and temporal features of pedestrians are extracted from the original data (acquired via multi-sensors), providing the basic pedestrian information for further analysis.

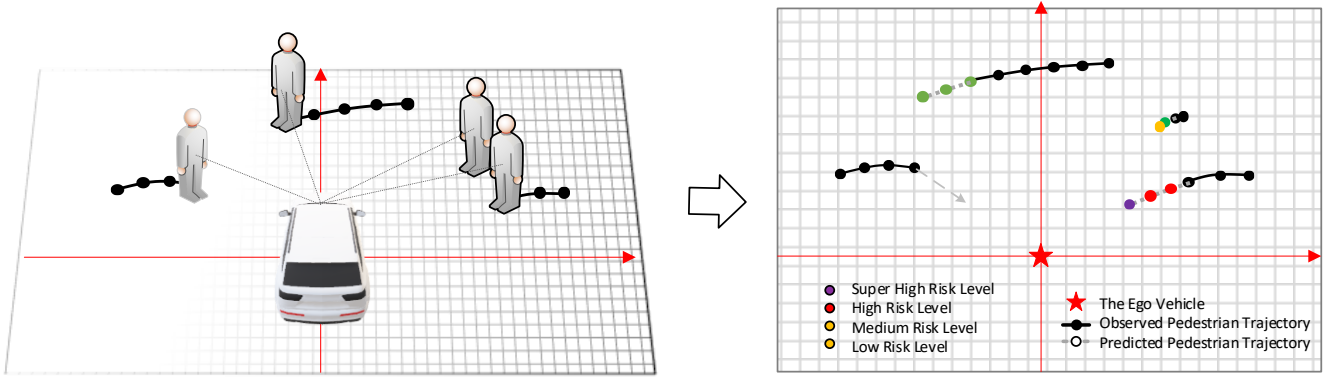

Fig 1.  A vehicle-pedestrian interaction scenario in the urban area.

When estimating a situation, the relative locations of surrounding pedestrians usually serve as a critical reference for drivers to compare with their own target places and decide on a safe time and place to move. Therefore, *Module 2* utilizes the collected vehicle-perspective data for training a pedestrian trajectory prediction model. Using the model, the location of pedestrians in the next several frames (also containing the ego vehicle's latent motion) can be predicted based on the formerly observed trajectories. Essentially, *Module 2* builds on data to provide a long-term estimation of the interactive intention of both the ego vehicle and the pedestrian. Thus, every predicted point in time and space yields a unique feature state, which belongs to a certain or several interactive intentions and risk patterns. To identify the risk patterns of the predicted trajectories with multiple spatiotemporal states, a risk level classifier is trained in *Module 3*.

*Module 3* utilizes another data-driven method to explore the risk pattern of pedestrians at a specific time and space and achieve risk level identification. The traditional method identifies the risk through a collision-based perspective, which offers a definite explanation for the risk-checking results based on a strict definition of collision. However, in real-world scenarios, pedestrians and vehicles are usually faced with a fuzzy interactive situation where no explicit conflict is present. Every change in spatial and temporal features of pedestrian constructs a unique interactive state, consequently influencing the risk state between the ego vehicle and pedestrian. Thus, a clustering method is utilized to explore the hidden risk patterns in the collected pedestrian data, especially regarding spatiotemporal features. The pattern does not necessarily provide a definite risk indicator or a collision probability but offers insights into the interactive intention between the ego vehicle and pedestrian and enables assessing the risk level of the current state. Once the risk pattern is discovered, a mapping relationship can be built between the pedestrian feature state and the risk level, which further serves to train a risk level classifier using the support vector machine (SVM) algorithm. The risk level classifier enables recognizing the risk level of new pedestrian feature states, which can be presented in a spatiotemporal form corresponding to the long-term predicted trajectory.

The overall system architecture is presented in Fig. 2.

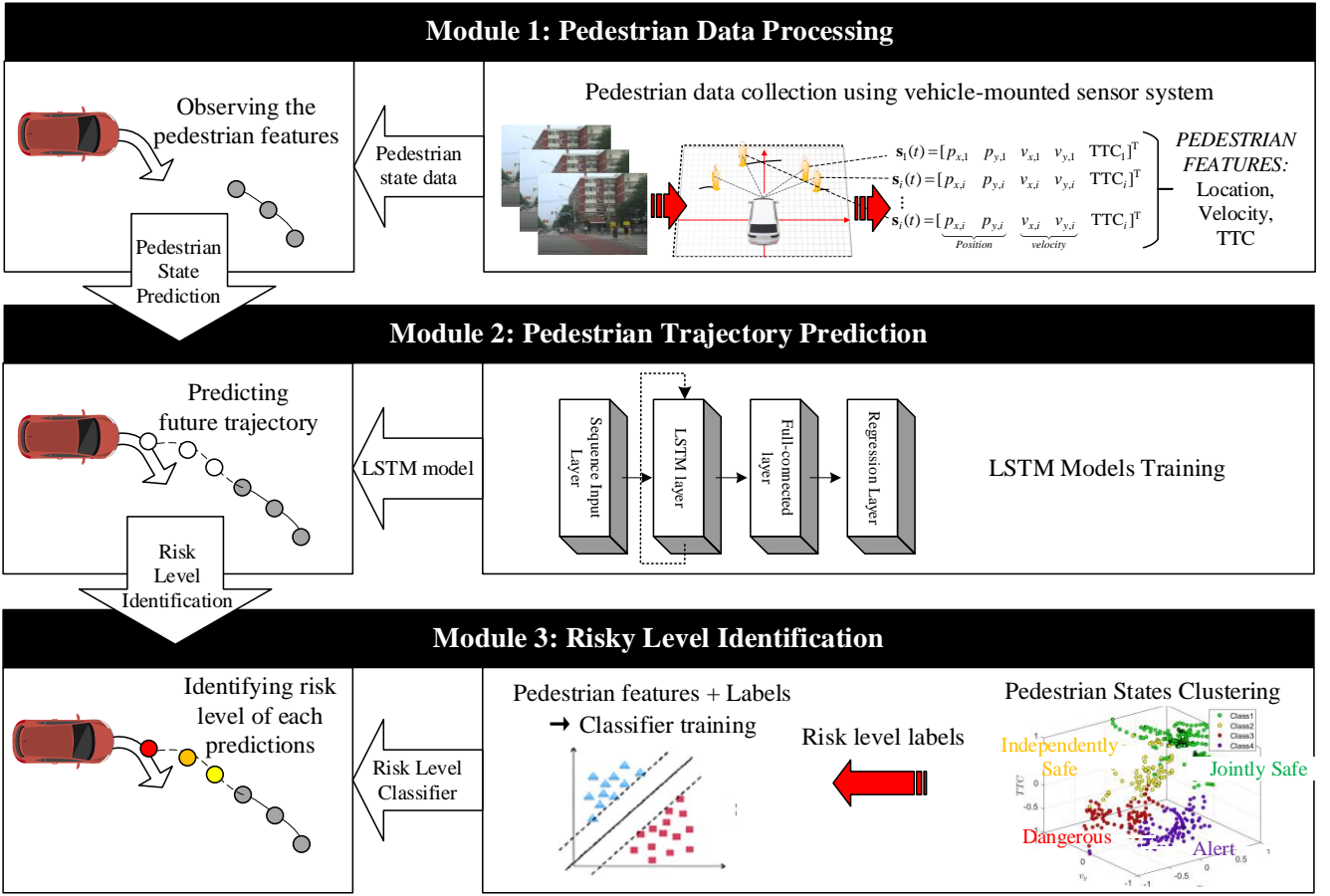

Fig 2. The overall system architecture. Module 1: The data are collected by vehicle-mounted sensors, and pedestrian features are extracted. Module 2: LSTM models are trained to build a time-series neural network. Such a network can infer the pedestrian movement pattern from observed features. Module 3: A hybrid clustering-classifying approach is proposed to learn human-vehicle risk patterns from vehicle-perspective data and predict the pedestrian risk level based on the learned risk patterns.

## 2.1. Module 1: Pedestrian Data Processing

A vehicle-mounted sensor system is utilized to collect 2D images and 3D point clouds of the surrounding area. By processing these data, pedestrian locations and other related features can be obtained. For a typical traffic scene of interest (shown in Fig. 1), at any time-instant $t$, the location vector of the $i^{th}$ pedestrian observed by the ego vehicle is denoted by $\mathbf{p}_i(t) = (x_i(t), y_i(t))$. Thus, for each pedestrian, the first instant (i.e., time step 1) is defined when the pedestrian is first observed by the ego vehicle. Now, the trajectory of the $i^{th}$ pedestrian from time 1 to $t$ can be described as:

$$\mathbf{P}_i(1,t) = \begin{bmatrix} \mathbf{p}_i(1) & \mathbf{p}_i(2) & ... & \mathbf{p}_i(t) \end{bmatrix}^T . \tag{1}$$

The location represents the spatial feature of the pedestrian. However, as discussed previously, it is insufficient to determine the risk using only location. Temporal features should also be considered. These features should offer sufficient information needed during the risk estimation process while driving. In this work, relative velocity and TTC are calculated as extended features used by the proceeding risk level identification module (i.e., *Module 2*). A clustering experiment in the following section studies how much these features reflect the risk.

Relative velocity $\mathbf{v}_i(t) = (v_x(t), v_y(t))$ is selected as a spatiotemporal feature. It captures whether the ego vehicle and the pedestrian are moving closer or drifting away from each other, considering the rate of change in the space between them in a short time. In this paper, $\mathbf{v}_i^{(t)}$ is calculated following the equations:

$$v_x(t) = (x_i(t) - x_i(t-1)) / \Delta t , \tag{2}$$

$$v_y(t) = (y_i(t) - y_i(t-1)) / \Delta t , \tag{3}$$

where $\Delta t$ is the time interval between two observations. While initializing a trajectory, the relative velocity is assumed to be 0.

Another spatiotemporal feature, TTC, is selected due to its capability to indicate the potential collision between the ego vehicle and pedestrian. As such, this metric is widely used as a risk indicator in many risk-related studies. Following [29], TTC is calculated in an approximate form by assuming a constant relative velocity until reaching a maximum time horizon (10 s in this work):

$$\text{TTC}_i(t) = \frac{\Delta p_i(t)}{\Delta v_i(t)}, \tag{4}$$

where $\Delta p_i(t) = \sqrt{x_i(t)^2 + y_i(t)^2}$ is the Euclidean distance between the pedestrian and the ego vehicle, and $\Delta v_i(t) = \mathbf{v}_i(t) \times \mathbf{p}_i(t)$ denotes the projection of $\mathbf{v}_i(t)$ in the direction of $\mathbf{p}_i(t)$ (see Fig. 3).

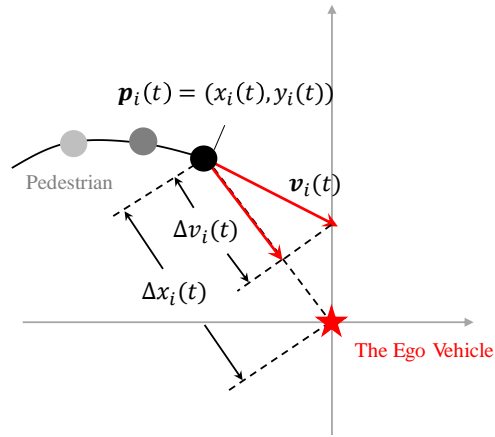

Fig 3.    Parameters used in calculating Time to Collision (TTC).

## 2.2. Module 2: Pedestrian Trajectory Prediction

Since solving complex dynamic functions bears an extensive computational cost, the proposed approach utilizes LSTM trained to predict pedestrian trajectories. LSTM is an artificial recurrent neural network, which has shown outstanding performance in sequence data processing due to its feedback mechanisms[17, 30]. Within this work, an LSTM network is constructed to predict pedestrian states. It comprises a sequence input layer, an LSTM layer, a fully connected layer, and a regression layer. The network input is the relative trajectory $\mathbf{p}_i(1,t) = \begin{bmatrix} \mathbf{p}_i(1) & \mathbf{p}_i(2) & ... & \mathbf{p}_i(t) \end{bmatrix}^T$, and the output is the expected future trajectory $\hat{\mathbf{p}}_i(t+1, t+T_{\text{pred}}) = \begin{bmatrix} \hat{\mathbf{p}}_i(t+1) & \hat{\mathbf{p}}_i(t+2) & ... & \hat{\mathbf{p}}_i(t+T_{\text{pred}}) \end{bmatrix}^T$ from time $t$ to $t+T_{\text{pred}}$, where $T_{\text{pred}}$ is maximized to capture long-term prediction. At any time-instant $t$, the LSTM layer comprises the following equations:

$$f(t) = \sigma(\mathbf{W}_f \cdot [\mathbf{h}(t\text{-}1), \ \mathbf{p}_i(t)] + \mathbf{b}_f), \tag{5}$$

$$i(t) = \sigma(\mathbf{W}_i \cdot [\mathbf{h}(t\text{-}1), \ \mathbf{p}_i(t)] + \mathbf{b}_i), \tag{6}$$

$$o(t) = \sigma(\mathbf{W}_o \cdot [\mathbf{h}(t\text{-}1), \ \mathbf{p}_i(t)] + \mathbf{b}_o), \tag{7}$$

$$\mathbf{c}(t) = f(t) \circ \mathbf{c}(t\text{-}1) + i(t) \circ \tanh(\mathbf{W}_c \cdot [\mathbf{h}(t\text{-}1), \ \mathbf{p}_i(t)] + \mathbf{b}_c), \tag{8}$$

$$\mathbf{h}(t) = o(t) \circ \tanh(\mathbf{c}(t)), \tag{9}$$

where $\sigma(\bullet)$ is the sigmoid activation function. Further, $\mathbf{W}_f$, $\mathbf{W}_i$, $\mathbf{W}_o$, and $\mathbf{W}_c$ denote the weights, whereas $\mathbf{b}_f$, $\mathbf{b}_i$, $\mathbf{b}_o$, and $\mathbf{b}_c$ stand for biases. Each LSTM layer delivers the cell state (denoted by $\mathbf{c}(t)$) to the next unit and outputs the hidden unit $\mathbf{h}(t)$. The objective is formalized as:

$$\arg\min_{\mathbf{W},\mathbf{b}} \frac{1}{n} \sum_{i=1}^{n} \sum_{t=1}^{m} (\mathbf{y}_i(t) - \tilde{\mathbf{y}}_i(t))^2, \tag{10}$$

where $y_i^{(t)}$ denotes the actual location of the $i^{th}$ pedestrian at time-instant $t$, $\tilde{y}_i^{(t)}$ denotes the predicted location, $n$ is the number of trajectories, and $m$ is the time duration of each trajectory. The LSTM network with optimal weights and biases is obtained by minimalizing the mean square error function.

For the observed pedestrian state $\mathbf{P}_i(1,t)$, the trained LSTM model aims to output the future trajectory of the $i^{th}$ pedestrian from time $t$ to $t+T_{pred}$, where $T_{pred}$ denotes a fixed prediction horizon. The output can be described as:

$$\hat{\mathbf{P}}_i(t+1,\ t+T_{pred}) = \begin{bmatrix} \hat{\mathbf{p}}_i(t+1) & \hat{\mathbf{p}}_i(t+2) & ... & \hat{\mathbf{p}}_i(t+T_{pred}) \end{bmatrix}^T. \tag{11}$$

From $\hat{\mathbf{p}}_i(t+1,\ t+T_{pred})$, other features, $\hat{\mathbf{v}}_i(t+1,\ t+T_{pred})$ and $\hat{\text{TTC}}_i(t+1,\ t+T_{pred})$, can be obtained using Eqs. (2), (3), and (4).

### 2.3. Module 3: Risk Level Identification

In various situations with different risk levels, pedestrians and vehicles interact differently, resulting in different pedestrian feature distributions. K-Means is a common clustering algorithm widely used due to its simplicity and efficiency. However, regular K-means use Euclidean distance to measure similarity between samples and might fail to distinguish data patterns when faced with non-convex data. Therefore, to capture the coupling effect of spatiotemporal factors on the pedestrian risk, a hybrid method integrating kernel principal component analysis and K-means clustering (KPCA-KMC) and spectral clustering (SC) are adopted for clustering the pedestrian feature data.

For every pedestrian at every timestep in a dataset, features are combined into a state vector $\mathbf{s}_i$. For each dataset, a set of pedestrian feature states $\mathbf{S} = \{\mathbf{s}_i(t) | i \in [1,n], t \in [1,T_n]\}$ is obtained, where $n$ denotes the total number of pedestrians in the dataset, and $T_n$ is the time length for each pedestrian. Using feature state set $\mathbf{S}$, the clustering algorithm can be applied to explore implicit risk patterns underlying different feature states. The discrepancy in the feature distributions is inspected and analyzed to distinguish the interactive intention and the risk level of different clusters.

As a data mapping tool, KPCA can be utilized for dimensionality reduction of high-dimensional input data. Furthermore, it may help determine the appropriate number of clusters ($K$) by analyzing the principal components [31]. In KPCA, input $\mathbf{S}$ is first projected onto a high-dimensional feature space $\phi$, thus enabling linear separation of the original data. Next, the conventional PCA is applied on $\phi(\mathbf{S})$, yielding a low-dimensional data projection $\hat{\phi}(\mathbf{S})$ on the principle components space, while preserving most of the variance in the original signal. After dimensionality reduction via KPCA, KMC is applied to partition observations $\hat{\phi}(\mathbf{S})$ into $K$ clusters (denoted by $\mathbf{C} = \{c_1, c_2, ... c_k\}$) by minimizing the within-cluster sum of squares. The objective of KMC is:

$$\arg\min_{\mathbf{C}} \sum_{k=1}^{K} \sum_{\hat{\phi}(\mathbf{s}_i) \in c_k} \| \hat{\phi}(\mathbf{s}_i) - \mu_k \|^2, \tag{12}$$

where $\mu_k$ is the centroid of the $k^{th}$ cluster.

SC is another effective clustering algorithm. It is based on graph theory and is considered for clustering the pedestrian feature data within this work due to its capability of dealing with high-dimensional data. SC relies on a spectral decomposition of the similarity matrix to reduce dimensionality before clustering the original data into fewer dimensions. Given dataset $\mathbf{S}$, an adjacency graph is first built using the $k$-nearest neighbors algorithm. The resulting graph consists of a set of nodes and the corresponding edges connecting neighboring points. The similarity matrix is then computed using the Gaussian kernel as a measure of similarity between different nodes. Using similarity matrix, graph Laplacian is calculated to perform eigenvalue decomposition. The eigenvalues of the Laplacian indicate the number of clusters, and the corresponding eigenvectors contain information on the appropriate segmenting of the graph nodes. $K$ can be obtained by analyzing the projection of the original data $\hat{\mathbf{S}}$ on the lower-dimensional manifold, ensuring minimal distortion. Finally, KMC is performed on $\hat{\mathbf{S}}$ to partition observations into $K$ clusters $\mathbf{C} = \{c_1, c_2, ... c_k\}$.

The input for both clustering methods is formulated as:

$$\mathbf{S} = [\mathbf{S}_1(1,T_1), ..., \mathbf{S}_i(1,T_i), ..., \mathbf{S}_n(1,T_n)]^T, \tag{13}$$

where $T_i$ represents the length of the $i^{th}$ pedestrian trajectory, and $n$ is the number of trajectories. The output, i.e., the obtained risk label set, is formulated as:

$$\mathbf{L} = [\mathbf{L}_1(1,T_1), ..., \mathbf{L}_i(1,T_i), ..., \mathbf{L}_n(1,T_n)]. \tag{14}$$

The criteria used when determining an appropriate number of clusters ($K$) and distinguishing a better clustering approach comprises three metrics, namely, the Akaike Information Criterion (AIC) and Bayesian Information Criterion (BIC) for determining $K$, and Silhouette for selecting the clustering approach. AIC is calculated as:

$$AIC = \sum_{k=1}^{K} \sum_{x^i \in c_k} |x^i - \mu^k|^2 + 2(K \cdot N), \quad (15)$$

where $x^i$ is the $i^{th}$ data point of the $k^{th}$ cluster, and $\mu^k$ is the $k^{th}$ cluster's centroid. Further, $K$ is the total number of clusters, and $N$ is the number of dimensions. The AIC metric evaluates the in-sample prediction error and penalizes the model complexity. A lower AIC value ensures reliable clustering results. Compared to AIC, BIC penalizes the model complexity more harshly by incorporating penalties for the number of data points. Formally,

$$BIC = \sum_{k=1}^{K} \sum_{x^i \in c_k} |x^i - \mu^k|^2 + \ln(M) \cdot (K \cdot N), \quad (16)$$

where $M$ is the total number of data points. By varying $K$ in KPCA-KMC and SC and searching for results with lower AIC and BIC values, the appropriate interval for $K$ can be determined.

The silhouette score is used to compare the clustering results of different approaches and determine the better one. The silhouette score is a succinct measure quantifying the similarities between an object and its assigned cluster with respect to other clusters [32]. It is calculated as:

$$a(i) = \frac{1}{|C_i| - 1} \sum_{j \in C_i, i \neq j} d(i, j), \quad (17)$$

$$b(i) = \min_{k \neq i} \frac{1}{|C_k|} \sum_{j \in C_k} d(i, j), \quad (18)$$

$$s(i) = \frac{b(i) - a(i)}{\max\{a(i), b(i)\}}, \quad (19)$$

where $a(i)$ denotes the mean distance between the $i^{th}$ data point and all other data points within the same cluster. In contrast, $b(i)$ denotes the minimum mean distance of the $i^{th}$ data point to all points in any other cluster (i.e., outside of its cluster). The Euclidean distance between $i^{th}$ and $j^{th}$ data points is denoted by $d(i, j)$. According to the definition, the silhouette score ranges from -1 to +1. The higher the silhouette score, the better the data point matches its cluster.

Once the pedestrian feature data is clustered into different clusters, the risk level classifier (i.e., SVM) can be trained. Then, the pedestrians' future risk levels can be identified by inputting the predicted pedestrian trajectories and the corresponding features into the trained risk level classifier. SVM is a widely used supervised learning algorithm that aims to find the maximum-margin hyperplane distinguishing the different examples. Thus, it can be used to classify new data points. Like KPCA, the kernel trick in SVM implicitly projects the original inputs into a high-dimensional feature space, potentially transforming the non-linear data into linearly separable ones. In this module, kernel SVM classifiers are trained for pedestrian risk level recognition. The classifier takes the pedestrian states $\mathbf{S} = [\mathbf{S}_1(1,T_1),...,\mathbf{S}_i(1,T_i),...,\mathbf{S}_n(1,T_n)]^T$ with risk labels $\mathbf{L} = [\mathbf{L}_1(1,T_1),...,\mathbf{L}_i(1,T_i),...,\mathbf{L}_n(1,T_n)]$ as inputs and outputs the risk level classifier that can match new state predictions to appropriate risk level categories.

Overall, *Module 2* provides the pedestrian trajectory prediction ($\hat{\mathbf{p}}_i(t+1, t+T_{pred})$) and the feature states ($\hat{\mathbf{S}}_i(t+1, t+T_{pred})$) for the $i^{th}$ pedestrian in the next $T_{pred}$ frames, whereas *Module 3* trains a classifier, using which a risk level sequence can be deduced from the predicted pedestrian's feature state. The output risk level sequence for each pedestrian can be represented as

$$\hat{\mathbf{L}}_i(t+1, t+T_{pred}) = \left[ \hat{\mathbf{l}}_i(t+1) \quad \hat{\mathbf{l}}_i(t+2) \quad ... \quad \hat{\mathbf{l}}_i(t+T_{pred}) \right]^T. \quad (20)$$

From a global perspective, the clustering of pedestrian feature states and the risk level analysis enable the proposed system to detect a change in risk pattern through a change in spatiotemporal features. Moreover, the risk identification is applied to a sequence of predicted features. It presents the risk in a spatiotemporal form, thus offering insights into the interaction pattern between the ego vehicle and pedestrians and identifying the (relatively long-term) risk level.

## 3. Experiments and Analysis

This section discusses the experimental details by first considering the data collection module, then the pedestrian trajectory prediction module, and finally, the risk level identification module.

*3.1. Pedestrian Data Processing Module*

The data collection module collects the data on pedestrian trajectories using a vehicle-mounted sensing system. The system utilized to obtain relative pedestrian trajectories in the "turning right" scenario belongs to the Intelligent Vehicle Research Center of Beijing Institute of Technology [33]. The data collection system contains a Velodyne LIDAR HDL-32E, an OxTs Inertial+ GNSS/INS suite, a Mako Camera, and an industrial computer (see Fig. 4).

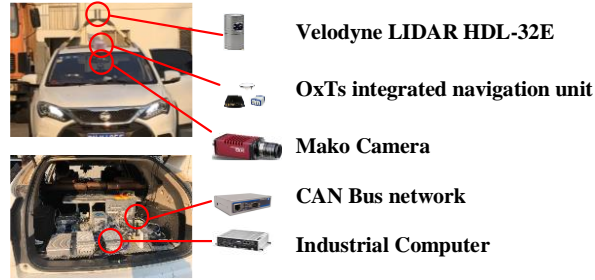

Fig 4. The vehicle-mounted sensing system utilized in data collection.

Velodyne LIDAR obtains point cloud information of surrounding obstacles, and Mako Camera collects images capturing the front vision of the vehicle. The onboard industrial computer is equipped for pedestrian detection and localization. The preloaded algorithm detects an individual pedestrian on a 2D image and provides relative coordinates by fusing the detection results with distance information from the point cloud. The origin of the relative coordinate system corresponds to the midpoint of the rear axle of the ego vehicle. The positive direction of the x-axis corresponds to the head orientation of the ego vehicle. Video and trajectory data are recorded at a frame rate of 6.5 fps.

As one of the busiest parts of the city road network, an urban intersection accommodates many pedestrian-vehicle encounters. Therefore, four signalized intersections on the Beijing Institute of Technology campus and West 3$^{rd}$ Ring Road in Beijing, China, were selected as the data collection locations. The driving route is shown in Fig. 5. The data recording was conducted during the daytime from 1 p.m. to 3 p.m. to ensure sufficient light and pedestrian flow. The data includes "turning right" and "going straight" scenarios. Two experienced drivers were selected to drive the vehicle. Finally, three hours of driving data were collected, and 84 pedestrian trajectories were extracted from the platform.

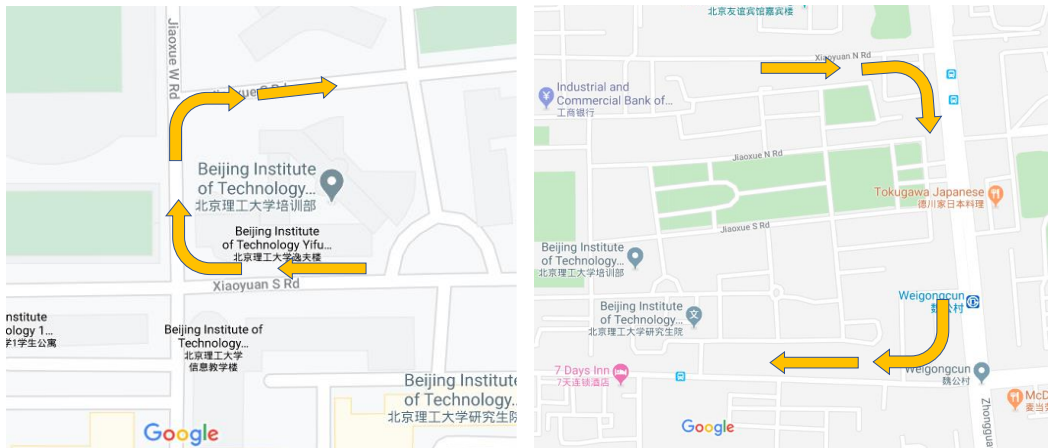

Fig 5. Driving route followed during the data collection.

The data collection was followed by preprocessing the pedestrian trajectories. Firstly, the trajectories in car-following scenes were manually removed due to a lack of direct interaction between the ego vehicle and a pedestrian. To reduce the random noise in the raw trajectories, the locally weighted scatter plot smooth (LOWESS) method [34] was utilized to smooth the original data. The data details are summarized in Table I.

TABLE I. DETAILS OF THE COLLECTED DATA

|  | **Campus** | **City** |
|---|---|---|
| Number of trajectories | 65 | 62 |
| Number of data points | 1611 | 1174 |
| Sampling frame fate (fps) | 6.5 | |
| Contained scenarios | Going straight, Turning right | |

The distributions of features (regarding position, velocity, and TTC) within the datasets are shown in Fig. 6. For each box, the central line indicates the median, and the bottom and top edges indicate the 25$^{th}$ and 75$^{th}$ percentiles, respectively. Here, $Pos_x$ and $Pos_y$ represent the relative pedestrian position along the $x$- and $y$-axis, and $Vel_x$ and $Vel_y$ denote the relative velocities.

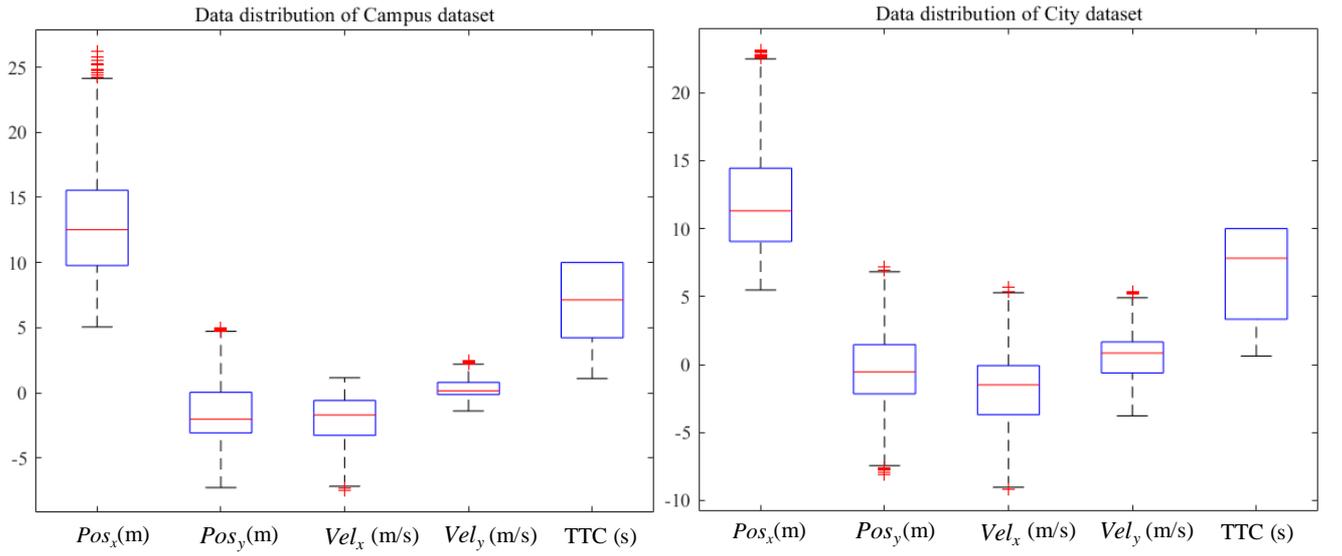

Fig 6.    Feature distributions in Campus and City datasets.

*3.2. Pedestrian trajectory prediction module*

The LSTM-based trajectory prediction model was trained within the pedestrian trajectory prediction module. An experiment was designed to find a broad prediction window while ensuring the prediction accuracy, thus ensuring a long-term pedestrian trajectory prediction. An observation window is set for trajectory prediction from frame 1 to frame $t$, and the prediction window ranged from frame $t+1$ to $t+T_{pred}$. For every trajectory, $t$ begins at the 4$^{th}$ frame and ends at the penultimate $T_{pred}^{th}$ frame. Five-fold Cross-Validation (CV) was employed to evaluate the system performance, reducing the variability and ensuring an accurate estimate of the model performance on a small dataset. For each dataset, ten repeats of the 5-fold CV were implemented. Finally, the training course was repeated eight times, with $T_{pred}$ varying from one to eight.

Following [17], the average displacement error (ADE) was used to evaluate the model performance. ADE denotes MSE averaged over all estimated locations on a trajectory and true locations. The results are shown in both the diagram and the table in Fig. 7.

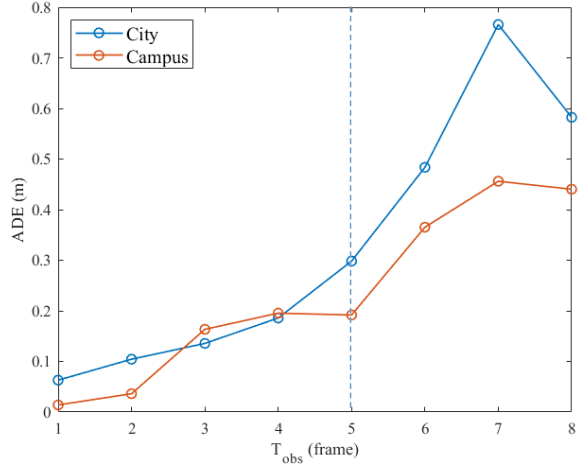

| $T_{pred}$ (frame) | ADE (m) of City data | ADE (m) of Campus data |
|---|---|---|
| 1 | 0.063018 | 0.013996 |
| 2 | 0.104408 | 0.036183 |
| 3 | 0.135605 | 0.163453 |
| 4 | 0.185838 | 0.195431 |
| **5** | **0.29831** | **0.191784** |
| 6 | 0.483684 | 0.365300 |
| 7 | 0.765843 | 0.456102 |
| 8 | 0.582529 | 0.440315 |

Fig 7.  Average displacement error (ADE) for different prediction windows. The selected prediction window size is indicated in bold.

Figure 7 shows that ADE grows as $T_{obs}$ increases for both datasets. However, ADE remains relatively low until $T_{pred} = 5$ (frames). When $T_{pred} > 5$ (frames), ADE sharply increases. Therefore, $T_{pred} = 5$ (frames) is selected as the prediction window for both datasets. Every pedestrian is observed from the 1st to the $t$-th frame, and then the trained LSTM model can predict their location in the next five frames, which can be represented as $\hat{\mathbf{p}}_i(t+1, t+T_{pred})$. With the predicted trajectory, other features of the pedestrian can be predicted (namely, velocity $\hat{\mathbf{v}}_i(t+1, t+T_{pred})$ and $\hat{TTC}_i(t+1, t+T_{pred})$) following Eqs. (2), (3), and (4).

*3.3. Risk level identification module I: Pedestrian features' clustering and classifier training*

As described in the previous section, the risk level identification is obtained by first applying the clustering methods to identify different risk patterns and then training the risk level classifiers on the clustering results. This process consists of four parts. First, the optimal number of clusters for each dataset is determined, and the most suitable algorithm (KPCA-KMC or SC) is selected. Next, the clustering results are compared using different pedestrian feature inputs. Then, the clustering results are discussed. Finally, the training of the risk level classifiers is presented.

*1) Selection of the optimal number of clusters and the clustering algorithm*

The number of clusters in the considered datasets is deduced by varying the $K$ values. For simplicity, all known pedestrian features (i.e., $\mathbf{s}_i(t) = [p_{x,i} \quad p_{y,i} \quad v_{x,i} \quad v_{y,i} \quad TTC_i]^T$) were used for clustering in this part. Figure 8 shows the performance of the two methods under different $K$ values on the City and Campus datasets. As seen in the left subfigure, the AIC and BIC curves of KPCA-KMC indicate that the optimal number of clusters for the City dataset is four. When $K = 4$, the BIC curve reaches its minimum, and AIC also reaches a relatively low level. Similar findings can be obtained for the Campus dataset, as Fig. 8 indicates $K = 4$ is the preferred selection for this case.

According to the results, the AIC and BIC of the SC method increase as K increases on both datasets. Unfortunately, such results enable little insight into how well the data is clustered using the SC method. Therefore, silhouette scores were used to distinguish the performance between the two clustering methods. Figure 9 shows the cluster silhouette scores and their mean value for KPCA-KMC and SC when $K = 4$. The mean silhouette scores of KPCA-KMC on the City and Campus datasets reach 0.51 and 0.42, respectively. In contrast, the silhouette scores are as low as 0.18 and -0.02 in the SC case. These results demonstrate that, while most data points have been clearly grouped using KPCA-KMC, SC fails to adequately separate the data. Therefore, KPCA-KMC is adopted for both datasets.

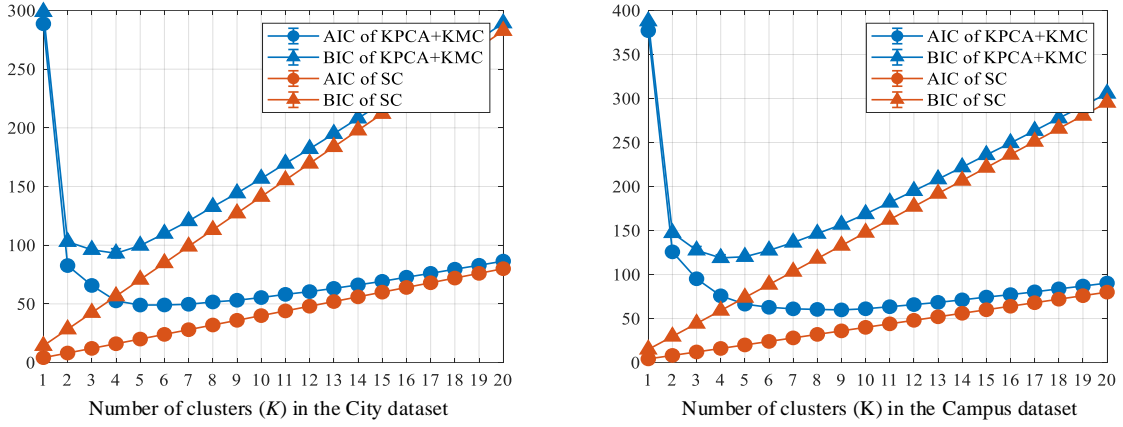

Fig 8.  AIC and BIC of KPCA-KMC and SC for different *K* values.

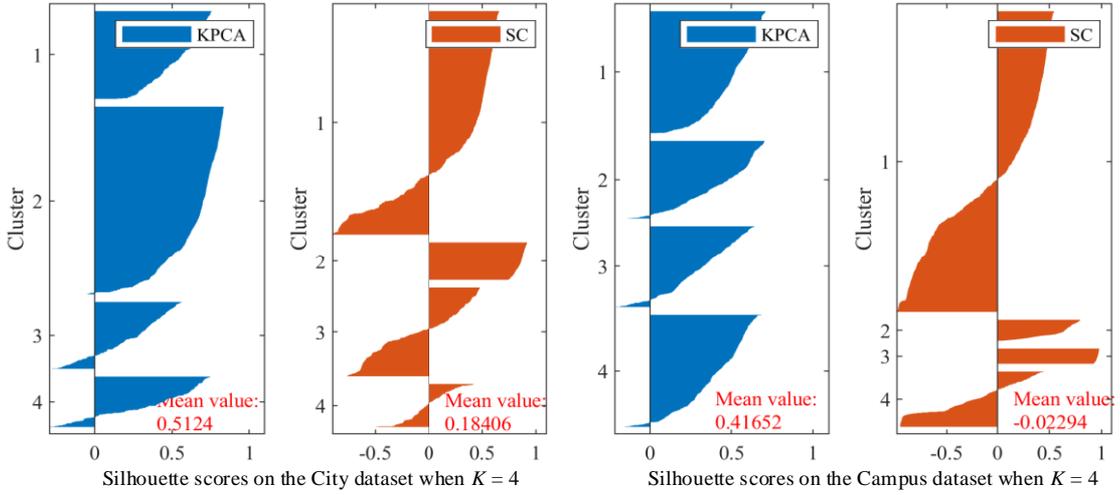

Fig 9.  The silhouette scores of the two datasets when *K* = 4.

### 2) *Comparison of the clustering results using different feature inputs*

This section studies how different features lead to different cluster results and risk level classifications. During clustering, the relative location, relative velocity, and TTC are used separately as the only pedestrian feature. In contrast, all the features are combined to form the pedestrian feature state. Different feature inputs for the $i^{\text{th}}$ pedestrian at time $t$ can be represented as:

$$\mathbf{s}_{p,i}(t) = [\underbrace{p_{x,i} \quad p_{y,i}}_{Location}]^{\text{T}}, \tag{21}$$

$$\mathbf{s}_{v,i}(t) = [\underbrace{v_{x,i} \quad v_{y,i}}_{velocity}]^{\text{T}}, \tag{22}$$

$$\mathbf{s}_{\text{TTC},i}(t) = [\text{TTC}_i]^{\text{T}}, \tag{23}$$

$$\mathbf{s}_{\text{all},i}(t) = [\underbrace{p_{x,i} \quad p_{y,i}}_{Location} \quad \underbrace{v_{x,i} \quad v_{y,i}}_{Velocity} \quad \text{TTC}_i]^{\text{T}}. \tag{24}$$

Using $\mathbf{s}_{p,i}(t)$, $\mathbf{s}_{v,i}(t)$, $\mathbf{s}_{\text{TTC},i}(t)$, and $\mathbf{s}_{\text{all},i}(t)$, the corresponding feature sets ($\mathbf{S}_p$, $\mathbf{S}_v$, $\mathbf{S}_{\text{TTC}}$, and $\mathbf{S}_{\text{all}}$) can be obtained for clustering. KPCA-KMC (*K* = 4) was applied on these four sets, and the clustering results for the City dataset are shown in Fig. 10. To obtain clustering results for $\mathbf{s}_{p,i}(t)$, $\mathbf{s}_{v,i}(t)$, and $\mathbf{s}_{\text{TTC},i}(t)$, pedestrians were divided only by their location/velocity/TTC. Figures 11 and 12 present the clustering results using different sets of pedestrian features.

Although clear boundaries are obtained in Fig. 10, the clustering results are not aligned. Pedestrians with similar location/velocity/TTC can differ in other feature dimensions and, thus, lead to a completely different risk level. These results indicate that the nearest pedestrian can be either the most dangerous or the safest when referring to the distinguishing criterion derived from clustering results using other features. Therefore, although each feature can indicate the risk to a certain extent (e.g., a closer relative location or higher velocity often indicates a riskier situation), using the features independently leaves uncertainties due to the lack of information, depending on how the vehicle and pedestrian interact with each other. On the other hand, the results in Fig. 11 and Fig. 12 show that when all features are used, the clustered pedestrian feature states present specific joint distributions that indicate a certain interactive intention between the ego vehicle and the pedestrian. Therefore, these feature states enable identifying different risk levels by including more critical information. Next, the clustering results using multiple features and the risk levels underlying different clusters are discussed.

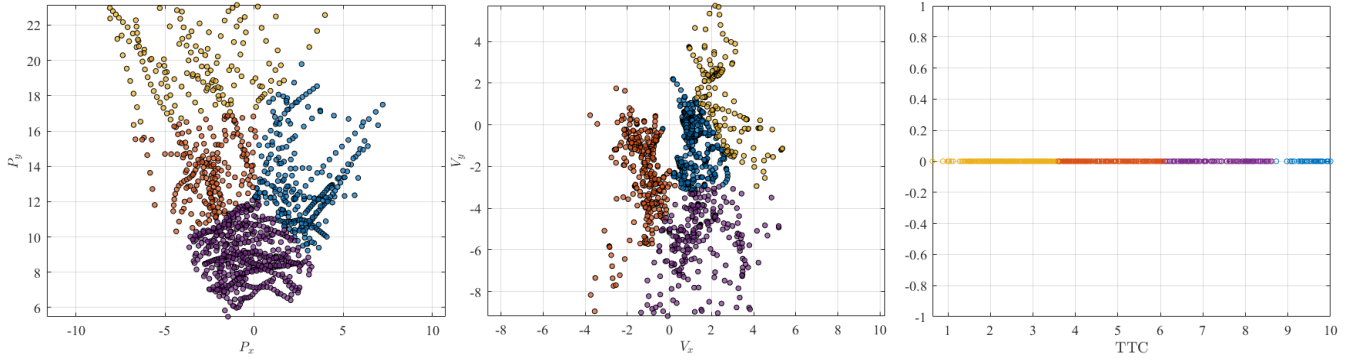

Fig 10.   Clustering results using only one feature, respectively location, velocity, and TTC.

*3) Clustering results discussions*

The clustering results obtained on the Campus dataset using KPCA+KMC with $K=4$ are shown in Fig. 11, whereas the results on the City dataset are seen in Fig. 12. To promote the analysis of the clustering results, the data are placed in two feature coordinate systems: $Pos_x - Pos_y - TTC$ coordinate system and $Vel_x - Vel_y - TTC$ coordinate system. Here, $Pos_x$ and $Pos_y$ represent the relative pedestrian position along the *x*- and *y*-axis, and $Vel_x$ and $Vel_y$ denote the relative velocities. Since the ego vehicle remains at the origin of the vehicle system, $(0,0,0)$ represents the ego vehicle in both coordinate systems (indicated with the red triangles in Fig. 11 and Fig. 12).

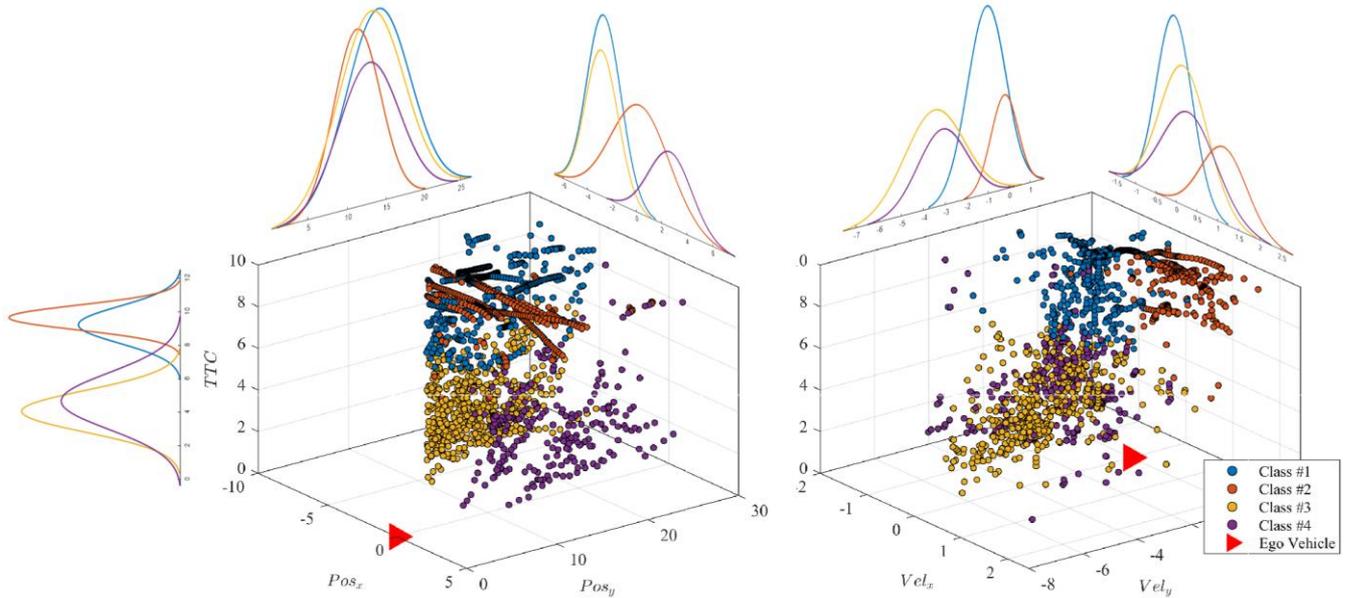

Fig 11.   The clustering results on the Campus dataset. Left: in $p_x - p_y - TTC$ coordinate system. Right: in $v_x - v_y - TTC$ coordinate system.

The clusters in Fig. 11 are analyzed as follows:

- Cluster #1 (blue): TTC is large, with a relatively large longitudinal distance from the ego vehicle and a low longitudinal speed. The lateral position mostly distributes to the left of the vehicle, and there is an overall tendency to continue moving left and drifting away from the ego vehicle. These characteristics indicate **no apparent conflict** between this pedestrian type and the vehicle currently. Therefore, the corresponding status is defined as **Independently Safe**.

- Cluster #2 (red): TTC is large, and the longitudinal position is close to the vehicle, yet the longitudinal speed is low. Meanwhile, the lateral position is widely distributed from left to right. Therefore, although a pedestrian is close to the ego vehicle, the pedestrian and vehicle pose **no danger** to each other. This case indicates the pedestrian is safely passing in front of the ego vehicle, under the situation that a right-of-way agreement has been reached. In this case, vehicles can **maintain their current operation**, e.g., stop and wait. Therefore, this status is defined as **Jointly Safe**, in which a consensus regarding the right-of-way has been reached.

- Cluster #3 (yellow): TTC is small. Further, there is a short longitudinal distance and left-leaning lateral position distribution, and the approaching speed is high, indicating that pedestrians (mostly from the left side) are approaching the ego vehicle. The situation is likely to turn into a conflict, requiring the vehicle to be **on high alert** and **prepared to avert to the right side or brake to avoid possible collisions from the left**. Therefore, this status is defined as **Dangerous**.

- Cluster #4 (purple): similar to Cluster #3, TTC is small, the longitudinal distance to the ego vehicle is short, and the approaching speed is high. However, most pedestrians are on the right side, requiring the ego vehicle to **be alert for pedestrians coming from the right-front**. Therefore, this status is defined as **Alert**.

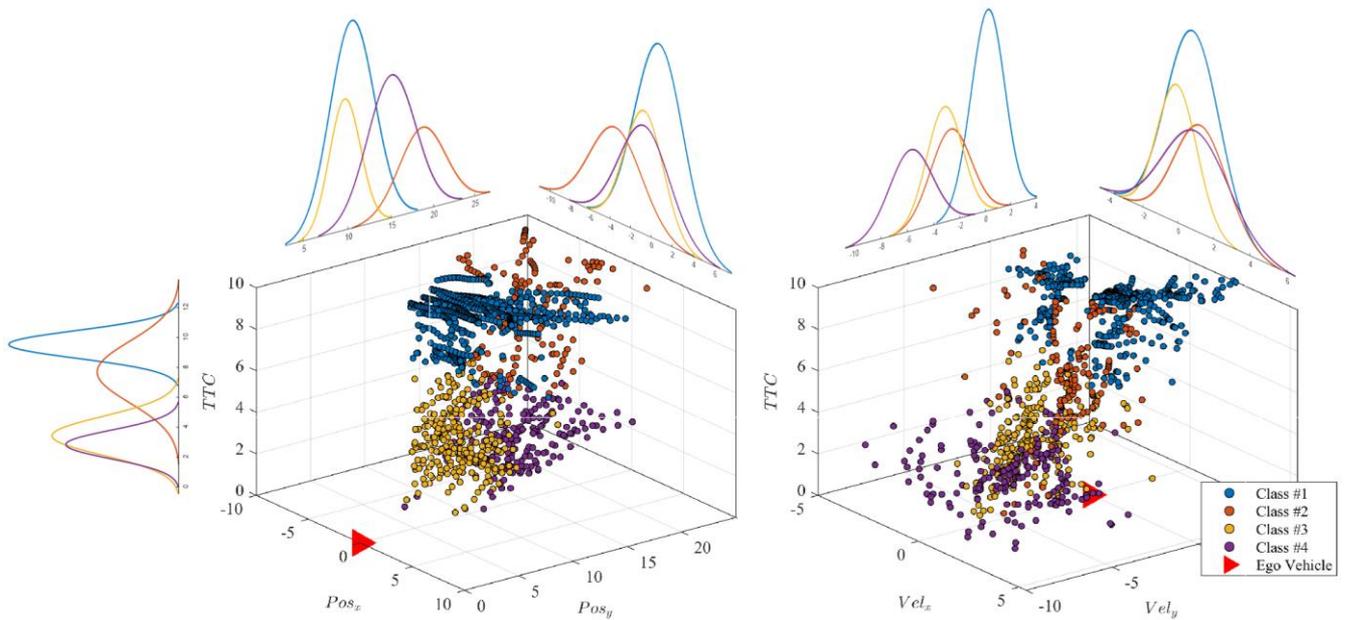

Fig 12. The clustering results on the City dataset. Left: in $p_x - p_y - TTC$ coordinate system. Right: in $v_x - v_y - TTC$ coordinate system.

The clusters in Fig. 12 are analyzed as follows:

- Cluster #1 (blue): Similar to Cluster #2 in the Campus dataset, the TTC is large, and there is a relatively short longitudinal distance to the ego vehicle and a low longitudinal speed. The lateral position is widely distributed from left to right, indicating pedestrian walking and crossing in front of the vehicle. In this case, vehicles can **maintain their current operation**, e.g., stop and wait. Therefore, this status is defined as **Jointly Safe**.

- Cluster #2 (red): TTC is large, the longitudinal distance to the ego vehicle is relatively large, and the approaching speed is moderate. The lateral position mostly distributes to the left of the vehicle, with a tendency of moving right, which indicates that pedestrians may be trying to cross in the far front of the vehicle. In this case, **slowing down or steering slightly** to a limited extent is recommended for the ego vehicle, but it is **unnecessary to take strong braking or steering measures**. Therefore, this status is defined as **Independently Safe**.

- Cluster #3 (yellow): TTC is small. The pedestrian is close to the vehicle with a medium approaching speed, indicating that pedestrians and the ego vehicle are very close yet still approaching each other. In this case, the ego vehicle should be **on high alert** and **avert to the opposite direction of the pedestrian or just brake to avoid possible collisions from the left**. Therefore, this status is defined as **Dangerous**.

- Cluster #4 (purple): TTC is small, the pedestrian is relatively far from the vehicle but with a high approaching speed, and shows crossing tendencies in the far front according to the lateral position and speed distribution. In this case, the vehicle should **be alert** and **be ready to make evasive moves in advance** to avoid the transition to a case similar to Cluster #3 (i.e., a dangerous situation). Therefore, this status is labeled as **Alert**.

*4) Risk level classifier training*

The presented discussion demonstrates that using multiple spatial and temporal features enables more accurate identification of the pedestrian risk level than when a single feature is utilized. The pedestrian risk level cannot be identified based on just one feature. To further verify this statement, two kinds of risk level classifiers are trained. One uses only the TTC feature as the input to identify the risk level (represented as $\Phi_{ttc}$). The other uses all features (namely, pedestrian location, velocity, and TTC) as input and is represented as $\Phi_{all}$. As the training data, $\Phi_{ttc}$ uses feature set $S_{ttc}$ with risk level label set $L_{ttc}$ (obtained from clustering results using TTC as input), whereas $\Phi_{all}$ employs $S_{All}$ with $L_{all}$ (derived from clustering results using all features as input).

Several SVM classifiers with different kernels were trained and evaluated using 5-fold CV for both datasets and both feature inputs. The classification performance of $\Phi_{all}$ is shown in Table II, whereas that of $\Phi_{ttc}$ is presented in Table III. The tables show that the classification performance of $\Phi_{ttc}$ is similar to that of $\Phi_{all}$, indicating a high classification accuracy of both input settings. The classifier with the highest accuracy is exported as the pedestrian risk level predictor for each dataset and both classifiers. The overall risk prediction performance is discussed in the next section.

TABLE II. THE SVM MODELS' PERFORMANCE OF $\Phi_{all}$

| Kernel type | Campus dataset | | City dataset | |
|---|---|---|---|---|
| | Accuracy | Prediction speed (obs/sec) | Accuracy | Prediction speed (obs/sec) |
| Linear kernel | 96.6% | ~6500 | 97.7% | ~5100 |
| Quadratic kernel | **98.0%** | ~8300 | 98.0% | ~7500 |
| Cubic kernel | 95.4% | ~4900 | 98.0% | ~8000 |
| Gaussian kernel | 97.9% | ~7900 | **98.4%** | **~8500** |

TABLE III. THE SVM MODELS' PERFORMANCE OF $\Phi_{ttc}$

| Kernel type | Campus dataset | | City dataset | |
|---|---|---|---|---|
| | Accuracy | Prediction speed (obs/sec) | Accuracy | Prediction speed (obs/sec) |
| Linear kernel | 97.8% | ~7600 | 96.6% | ~6100 |
| Quadratic kernel | 98.1% | ~8300 | 97.6% | ~7200 |
| Cubic kernel | 97.0% | ~4500 | **99.0%** | **~8500** |
| Gaussian kernel | **98.9%** | ~8200 | 98.8% | ~8100 |

*3.4. Risk level identification module II: Testing the pedestrian risk level prediction*

Using feature predictions $\hat{\mathbf{p}}_i(t+1, t+T_{pred})$, $\hat{\mathbf{v}}_i(t+1, t+T_{pred})$, and $\hat{TTC}_i(t+1, t+T_{pred})$ from the trajectory prediction module, the trained classifiers can predict future risk level $\hat{\mathbf{L}}_i(t+1, t+T_{pred})$. Two typical situations are selected and shown in Fig. 13 to further discuss the predictions of the classifiers on the two datasets. The original trajectory corresponds to the grey line, and the predicted trajectory with risk level labels corresponds to the dotted colored line, where color indicates different risk levels. The main plot shows the risk level identified by considering all features, whereas the grey box contains the classification result based on TTC only.

In the Campus example, the ego vehicle is going straight when it detects a relatively distant pedestrian showing an intention to cross. As the vehicle continues moving forward, the pedestrian makes the cross. Let *T* denote a specific time instant. It can be noticed that the given risk level predictions changed from Red (Jointly safe) to Blue (Independently Safe) around *T* = 6, from Blue (Independently Safe) to Purple (Alert) around *T* = 11, and from Purple (Alert) to Red (Jointly Safe) around *T* = 16. The prediction gives the ego vehicle a broad hint of the future risk and correctly identifies the probing interaction between the ego vehicle and the pedestrian. Consequently, the ego vehicle obtains practical suggestions each time instant (for example, the turning alert at $T = 11$). In contrast, the classifier that uses only TTC as input ignores the changes in the pedestrians' location and velocity as time passes. It only gives a rough estimation of whether TTC is getting higher or lower. Such a classifier fails to predict the real risk level and distinguish various situations as the spatiotemporal relationship changes.

In the City example, the ego vehicle is turning right when it firstly notices a pedestrian approaching from the right side. As the ego vehicle continues turning right, the pedestrian is approaching and shows no intention to stop. Finally, the ego vehicle makes a concession, and the pedestrian crosses the road in front of the ego vehicle. The given predictions again undergo three stages, switching from Red (Independently Safe) to Purple (Alert) around $T = 4$, from Purple (Alert) to Yellow (Dangerous) around $T = 11$, and from Yellow (Dangerous) to Blue (Jointly Safe) around $T = 21$. Similar to the Campus example, the prediction successfully identifies the implicit interaction between the ego vehicle and the pedestrian and provides practical suggestions, like taking evasive actions at $T = 11$. In this case, the TTC classifier shows weak flexibility and remains inaccurate in identifying the risk level throughout the process.

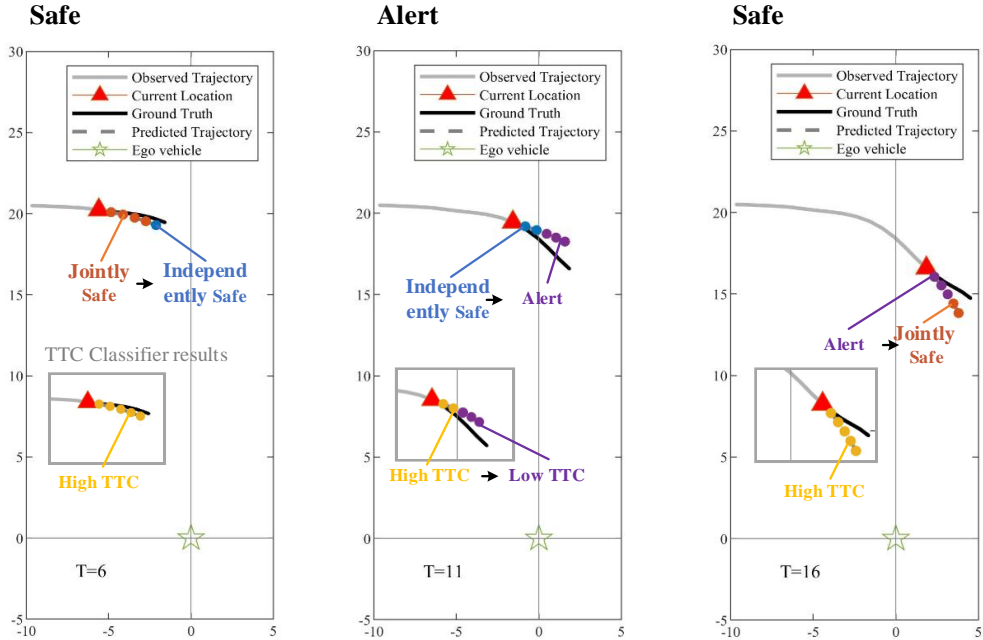

a) Campus Example

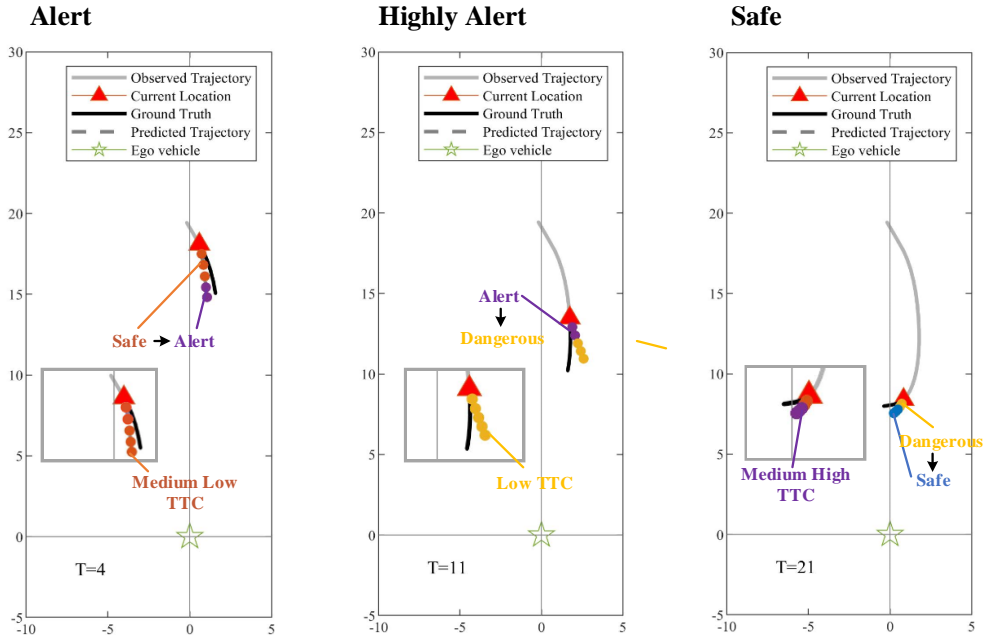

b) City Example

Fig 13. Examples of the risk level prediction.

The trained LSTM model provided reliable pedestrian state and risk level predictions in the two cases by considering the implicit interaction between the ego vehicle and the pedestrian as their spatiotemporal relationship changed over time. The risk level prediction enables the ego vehicle to anticipate the risks and their possible effects under the current situation while providing valuable insights for the planning process.

The performance of the system was assessed by comparing the predicted risk level $\hat{\mathbf{L}}_i(t+1, t+T_{pred})$ to the actual risk level $\mathbf{L}_i(t+1, t+T_{pred})$ for every pedestrian in both datasets. The confusion matrices are shown in Fig. 14, demonstrating that the model reaches the overall accuracy of 83.7% on the Campus dataset and 87.3% on the City dataset. Thus, the model exhibits a good performance when predicting the pedestrian risk level. Since the proposed system aims to enhance driving security, the error rate when predicting "Alert" and "Dangerous" should be paid special attention. In particular, one should inspect the cases where a risky situation is mistaken for a safe one, thus potentially leading to a reckless and catastrophic event. The Target Class bar (bottom grey bar) shows that the error rates for "Dangerous" (i.e., Class #3) and "Alert" (i.e., Class #4) classes are 2.5% and 9.7% on the Campus dataset, and 19.9% and 7.5% on the City dataset. The error rate on the Campus scenario is lower than that on the City dataset, which might be attributed to the simpler interaction situation in the Campus case. The higher error rate in the urban scenario is especially evident in the "Dangerous" class, with 19.9% inaccurate predictions. The results demonstrate that the system tends to identify "Dangerous" cases as "Jointly Safe" (3.0%) or "Alert" (1.9%), mistaking a situation where a vehicle needs to act immediately for a safe situation or a situation requiring caution. This behavior necessarily causes several safety problems. However, as demonstrated earlier, the system does not yield a risk level prediction at a particular moment in time but constructs a sequence of future risk levels associated with time and space. By reviewing and judging the spatiotemporal risk levels over a period, the interference of such misclassification can be eliminated to a certain extent. Nevertheless, the proposed system still suffers prediction errors due to our limited understanding of human-vehicle interaction. Additional information related to human-vehicle interaction (e.g., pedestrian skeletons or hand gestures) will be added to this system as a basis for risk assessment in the future. Overall, the experiment results demonstrate that the proposed system successfully identifies the spatiotemporal risk levels of pedestrians on both datasets. Thus, the system accurately assesses different interactive intentions.

**Confusion Matrix on the Campus dataset**

| Output Class \ Target Class | Independently Safe | Jointly Safe | Dangerous | Alert | |
|---|---|---|---|---|---|
| Independently Safe | 1297 / 15.7% | 52 / 0.6% | 21 / 0.3% | 0 / 0.0% | 94.7% / 5.3% |
| Jointly Safe | 108 / 1.3% | 1640 / 19.9% | 20 / 0.2% | 57 / 0.7% | 89.9% / 10.1% |
| Dangerous | 741 / 9.0% | 22 / 0.3% | 2557 / 31.0% | 95 / 1.2% | 74.9% / 25.1% |
| Alert | 75 / 0.9% | 133 / 1.6% | 24 / 0.3% | 1418 / 17.2% | 85.9% / 14.1% |
| | 58.4% / 41.6% | 88.8% / 11.2% | 97.5% / 2.5% | 90.3% / 9.7% | **83.7% / 16.3%** |

**Confusion Matrix on the City dataset**

| Output Class \ Target Class | Jointly Safe | Independently Safe | Dangerous | Alert | |
|---|---|---|---|---|---|
| Jointly Safe | 2339 / 44.6% | 29 / 0.6% | 159 / 3.0% | 2 / 0.0% | 92.5% / 7.5% |
| Independently Safe | 2 / 0.0% | 277 / 5.3% | 12 / 0.2% | 29 / 0.6% | 86.6% / 13.4% |
| Dangerous | 142 / 2.7% | 12 / 0.2% | 1094 / 20.9% | 39 / 0.7% | 85.0% / 15.0% |
| Alert | 22 / 0.4% | 119 / 2.3% | 100 / 1.9% | 868 / 16.5% | 78.3% / 21.7% |
| | 93.4% / 6.6% | 63.4% / 36.6% | 80.1% / 19.9% | 92.5% / 7.5% | **87.3% / 12.7%** |

Fig 14.    The risk level prediction model's confusion matrices for the two datasets.

## 4. Conclusions

In this paper, a novel PRLP system is proposed to predict the spatiotemporal pedestrian risk level. The system contains three modules: pedestrian data preprocessing, pedestrian trajectory prediction, and risk level identification. First, the vehicle-perspective data is collected. Two datasets containing the spatiotemporal features of pedestrians (namely, the Campus and the City datasets) were obtained within the data preprocessing module. The collected vehicle-perspective pedestrian data were then used to train the trajectory prediction module for predicting the location and other spatiotemporal features of pedestrians in the subsequent five frames. The risk level identification module revealed four risk levels discovered among the pedestrian feature data in both datasets. These risk levels were labeled Jointly Safe, Independently Safe, Dangerous, and Alert. Based on the learned risk patterns, risk level classifiers were trained to assess the risk level of predicted pedestrian feature states. The final results showed that the proposed system effectively identifies the interactive intentions between the ego vehicle and the pedestrian, predicting the spatiotemporal risk level of pedestrians with high accuracy.

The vehicle-perspective data collected in this work offer direct pedestrian-vehicle interaction information, sparing complex modeling work based on road-perspective data. Moreover, through entirely data-driven methods, the proposed system can extract the motion and risk patterns of a pedestrian from the data, predicting the risk level of the pedestrian in a spatiotemporal manner. The predicted risk level is not an instantaneous risk indicator but an integrated representation of risk, providing insights into the interaction pattern between the ego vehicle and the pedestrian which changes through time and space. This work can be used to develop an onboard warning system to increase drivers' awareness of pedestrian risk. Furthermore, it can serve as a valuable reference for a decision-making module in autonomous driving systems.

In the future, more pedestrian feature data will be added, and modifications will be made on both the pedestrian trajectory prediction module and the risk level identification module to enable longer-term predictions and yield more interpretable risk patterns. Building on the utilized data-driven paradigm, the proposed system will be extended to include driving scenarios containing more types of road users (e.g., cyclists and other vehicles), thus further advancing the ability of intelligent driving systems to assess risks in complex driving scenes.